\documentclass[11pt,letterpaper]{article}
\usepackage[margin=0.9in,headheight=14pt]{geometry}
\usepackage[T1]{fontenc}
\usepackage[utf8]{inputenc}
\usepackage{lmodern,microtype}
\usepackage{amsmath,amssymb}
\usepackage{graphicx,booktabs,tabularx,array}
\usepackage{enumitem,listings,xcolor,caption,float}
\usepackage[most]{tcolorbox}
\usepackage{fancyhdr,needspace,placeins,etoolbox}
\usepackage[numbers,sort&compress]{natbib}
\usepackage{xurl}
\usepackage{hyperref}
\definecolor{ink}{HTML}{213547}
\definecolor{academicblue}{HTML}{0066CC}
\definecolor{referenceteal}{HTML}{008C8C}
\definecolor{linkcopper}{HTML}{C05000}
\definecolor{pale}{HTML}{F3F6F8}
\definecolor{teal}{HTML}{007A79}
\hypersetup{colorlinks=true,linkcolor=referenceteal,citecolor=academicblue,urlcolor=linkcopper,filecolor=linkcopper,linktoc=all,breaklinks=true,pdftitle={Tutorial: Vision-Based Formation Control with Differential-Drive Robots},pdfauthor={Christian Marowski and Byung-Cheol Min}}
\setlist{nosep,leftmargin=1.5em,topsep=4pt}
\newcolumntype{Y}{>{\raggedright\arraybackslash}X}
\lstdefinestyle{python}{language=Python,basicstyle=\ttfamily\small,keywordstyle=\color{academicblue}\bfseries,commentstyle=\color{teal},stringstyle=\color{brown},showstringspaces=false,columns=fullflexible,keepspaces=true,breaklines=true,frame=single,rulecolor=\color{black!15},backgroundcolor=\color{pale},xleftmargin=4pt,xrightmargin=4pt,aboveskip=8pt,belowskip=8pt,tabsize=4}
\newtcolorbox{tutorialnote}[1]{enhanced,breakable,colback=pale,colframe=academicblue!40!white,boxrule=.5pt,arc=1pt,left=8pt,right=8pt,top=5pt,bottom=5pt,title={#1},colbacktitle=academicblue!8!white,coltitle=ink,fonttitle=\bfseries}
\newcommand{\norm}[1]{\left\lVert#1\right\rVert}
\newcommand{\code}[1]{\texttt{\detokenize{#1}}}
\newcommand{\repo}{\url{https://github.com/SMARTlab-IU/formation-control-tutorial}}

\title{\textbf{Tutorial: Vision-Based Formation Control\\with Differential-Drive Robots}}
\author{Christian Marowski and Byung-Cheol Min\\[3pt]\normalsize SMART Laboratory\\\normalsize Luddy School of Informatics, Computing, and Engineering\\\normalsize Indiana University, Bloomington, IN 47405, USA\\\normalsize\texttt{\{cmarows,minb\}@iu.edu}}
\date{}

\begin{document}
\maketitle

\vspace{-20pt}
\begin{abstract}
\noindent This tutorial presents a step-by-step introduction to multi-robot formation control using differential-drive robots, an overhead camera, and Python. ArUco markers are used to estimate robot positions and orientations, while desired pairwise displacement vectors define a circular formation. By comparing measured and desired relative positions, the controller computes movement vectors that are converted into differential-drive wheel commands, with basic proximity-sensor-based obstacle avoidance. The tutorial guides readers through hardware and software setup, robot pose estimation, formation geometry, feedback control, system integration, and formation-error evaluation. Motivated by the need to bridge the gap between theoretical understanding and practical implementation in robotics education, we emphasize how mathematical concepts are translated into coordinated behaviors on physical robots. Although the implementation uses a specific robotic platform, the underlying perception and formation-control principles can be adapted to other differential-drive robots and simulation environments. The accompanying open-source GitHub repository provides the Python implementation, setup instructions, and demonstrations to support hands-on learning and further exploration of multi-robot systems.
\end{abstract}

\smallskip
\vspace{-6pt}
\textbf{~~~~~~~~GitHub Repository:} \repo

\section{Introduction}
\label{sec:introduction}

Formation control enables a team of robots to achieve and maintain a desired spatial arrangement through coordinated motion. Unlike classical point-convergence and rendezvous problems, in which robots move toward a common location~\citep{ando1999rendezvous,lin2007rendezvous}, formation control seeks to establish specified spatial relationships among robots. This capability is fundamental to multi-robot coordination and provides a useful setting for studying how individual robot behaviors contribute to a collective objective. However, translating formation-control principles into physical robot motion requires more than understanding the underlying mathematical models. It also involves integrating sensing, coordinate transformations, feedback calculations, and motion control into a functioning robotic system.

Formation control has been extensively studied through various approaches, including behavior-based coordination~\citep{balch1998formation}, control of nonholonomic mobile robots~\citep{desai2001modeling}, and formation-constrained multi-agent control~\citep{egerstedt2001formation}. A comprehensive survey by Oh et al.~\citep{oh2015survey} categorizes major formation-control strategies and their underlying principles. Related research has also explored beacon-based distributed structure formation~\citep{mina2023beacon}, multipoint rendezvous~\citep{parasuraman2020multipoint}, bearing-aided rendezvous using network topology~\citep{luo2019bearing}, and communication-aware robot coordination, including directional-antenna-based leader--follower relaying~\citep{min2018directional} and consensus using the direction of arrival of wireless signals~\citep{parasuraman2019doa}. These studies illustrate different approaches to multi-robot coordination, sensing, and communication. While such methods provide important theoretical and algorithmic foundations, implementing even a relatively simple formation controller on physical robots introduces practical challenges that are valuable for students to understand.

A primary motivation for developing this tutorial is to make these fundamental concepts more accessible through hands-on learning. In robotics education, students often encounter mathematical models and control algorithms before having opportunities to observe how they behave on physical systems. We believe that connecting theory with implementation is an essential part of learning robotics, particularly in multi-robot systems, where sensing, coordination, and individual robot motion must work together. Accordingly, this tutorial is designed to guide students from the basic geometric principles of formation control to their implementation and demonstration on physical robots, helping them develop both conceptual understanding and practical experience.

\begin{figure}[tbp]
    \centering
    \includegraphics[width=.92\textwidth]{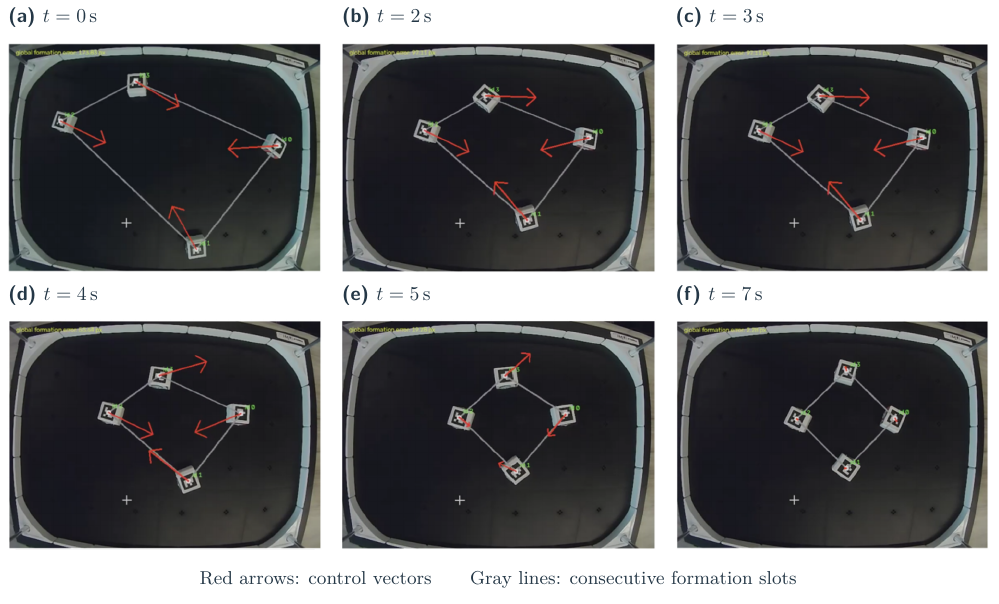}
        \vspace{-3pt}
    \caption{Formation acquisition by four robots in one experimental run at $t=0,2,3,4,5,7$ seconds. Red arrows show control vectors; gray lines connect robots in consecutive formation slots. The sequence provides a qualitative illustration of the process. The frame annotations label the four robots 0--3; the example configuration uses marker IDs 0, 1, 2, and 3.}
    \label{fig:sequence}
    \vspace{-3pt}
\end{figure}

This tutorial presents a step-by-step implementation of vision-based formation control using a team of differential-drive robots (e.g., Robomation's HamsterS), an overhead camera, and Python. ArUco markers are used to identify individual robots and estimate their positions and orientations in image coordinates. Each robot is assigned a fixed slot among equally spaced points on a circle, and the desired formation is represented through pairwise displacement vectors. By comparing the measured relative positions with the desired displacements, the controller computes a movement vector for each robot, which is subsequently translated into turning and forward wheel commands. Because the formation is defined through relative displacements rather than fixed global destinations, the robots can establish the desired arrangement at different locations within the camera's field of view. Figure~\ref{fig:sequence} illustrates the formation acquisition process using four physical robots.

Throughout the tutorial, we emphasize the relationship between mathematical formulations and their practical implementation. Readers will learn how to estimate robot poses from camera images, construct desired formation geometries, calculate displacement errors, and convert control vectors into differential-drive motion. We also address practical considerations such as consistent coordinate conventions, robot identification, controller parameters, obstacle responses, and formation-error evaluation. The accompanying GitHub repository provides the Python implementation, setup instructions, and demonstrations~\citep{tutorialrepo}, enabling readers to reproduce the experiments and further explore the concepts presented here. We hope this tutorial provides an accessible starting point for students and researchers interested in robotics, particularly multi-robot systems, and encourages further learning and experimentation in robot coordination and control.

\section{System Overview}
\label{sec:overview}

Our system consists of a team of differential-drive robots, a stationary overhead USB camera, and a central computer running Python. Each robot carries a unique ArUco marker, which OpenCV uses to estimate its position and heading in image coordinates. The computer compares the measured relative positions with the desired pairwise displacements of a circular formation and computes a control vector for each robot. These vectors are converted into differential-drive wheel commands and transmitted through RoboID. The process repeats with each new camera frame, forming a continuous perception-and-control loop.

\begin{figure}[htbp]
\centering
\vspace{5pt}
\includegraphics[width=\textwidth]{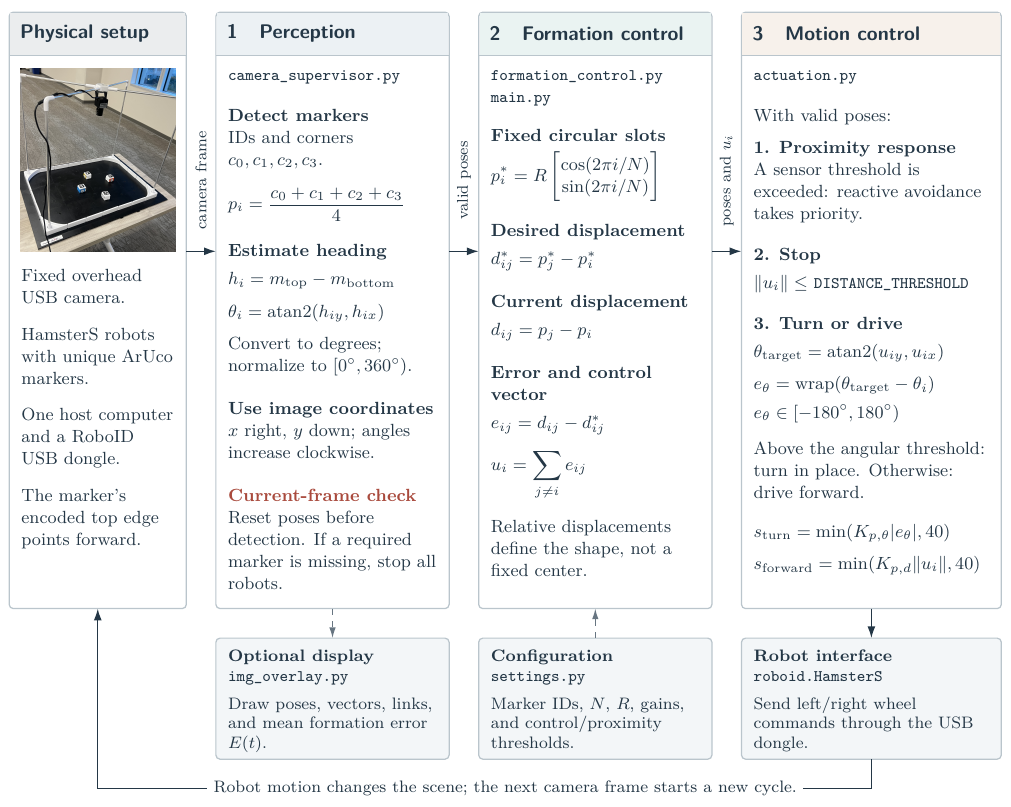}
\caption{System architecture. A fixed overhead camera observes the marked robots; the central computer estimates poses, computes pairwise formation errors, and sends wheel commands through RoboID. The configuration supplies marker IDs, slot geometry, gains, and thresholds. The dashed display block is optional.}
\label{fig:system_architecture}
\end{figure}

The implementation is \textit{centralized}: a single computer processes all observations and computes commands for the entire team. If any required marker is not detected, all robots stop until valid observations are restored. Onboard proximity sensors also provide basic reactive obstacle avoidance. Figure~\ref{fig:system_architecture} illustrates the overall system architecture, while Figure~\ref{fig:architecture} summarizes the control decisions within each iteration.

Unlike our centralized implementation, Fredslund and Matari{\'c}~\citep{fredslund2002general} investigate formation control using local sensing and minimal communication, while Lawton et al.~\citep{lawton2003decentralized} study decentralized formation maneuvers.

\subsection{Project structure}
The program is organized into \code{main.py}, \code{settings.py}, and four modules in \code{utils/}. Table~\ref{tab:files} summarizes their roles.

\begin{table}[htbp]
\centering\small
\caption{Organization of the Python implementation.}
\label{tab:files}
\begin{tabularx}{\textwidth}{@{}p{.37\textwidth}Y@{}}
\toprule
File & Role\\
\midrule
\code{main.py} & Initializes robots and executes the main control loop.\\
\code{settings.py} & Stores marker IDs, formation parameters, gains, and thresholds.\\
\code{utils/camera_supervisor.py} & Detects markers and updates robot positions and orientations.\\
\code{utils/formation_control.py} & Constructs formation slots and desired relative displacements.\\
\code{utils/actuation.py} & Converts a control vector into wheel commands and checks proximity sensors.\\
\code{utils/img_overlay.py} & Provides optional pose, vector, and formation-error visualization.\\
\bottomrule
\end{tabularx}
\end{table}

\FloatBarrier

\section{Hardware Setup}
\label{sec:hardware}
\subsection{Requirements}
The example uses two or more HamsterS differential-drive robots, a fixed overhead USB camera, one distinct printed ArUco marker per robot, and a computer capable of running Python. We recommend a camera capable of 1080p resolution. Other differential-drive robots can use the formation calculations and marker-based pose-estimation approach, but their connection libraries, sensor calls, and wheel-command units may differ.

\subsection{Generating and mounting markers}
ArUco markers are square fiducial markers that can be identified from their encoded patterns~\citep{garridojurado2014aruco}. Use the same dictionary during generation and detection. The example configuration uses \code{DICT_7X7_100}. The following excerpt generates marker ID 0, the first marker in that configuration:
\begin{lstlisting}
import cv2

dictionary = cv2.aruco.getPredefinedDictionary(
    cv2.aruco.DICT_7X7_100
)
marker = cv2.aruco.generateImageMarker(
    dictionary, 0, 500
)
cv2.imwrite("marker_0.png", marker)
\end{lstlisting}
Generate a different ID for each robot. Print the markers flat, retain a clear white margin, and attach them securely so that they cannot rotate relative to the robot. See the OpenCV documentation for the marker-generation and detection APIs~\citep{opencv2026aruco}.

\textbf{Mounting direction matters.} The top edge of the generated marker image must point toward the robot's front. The orientation calculation in Section~\ref{sec:pose} uses this convention. The marker center represents the robot center only approximately unless the marker is mounted centrally.

\vspace{-1.5pt}
\subsection{Camera placement}
\vspace{-1.5pt}
Secure the camera above the workspace, approximately perpendicular to the floor. Keep every marker within view and maintain adequate, consistent lighting. Motion blur and autofocus changes can reduce detection reliability; fixed focus can help. The camera should remain stationary during operation.

\vspace{-1pt}
\section{Software Setup and Configuration}
\vspace{-1pt}
\label{sec:software}
Our implementation uses Python 3.10 or newer, NumPy~\citep{harris2020numpy} for array calculations, OpenCV~\citep{bradski2000opencv} with the ArUco module for image processing, and RoboID for communication with HamsterS robots. The \code{RobotAgent} annotations use the \code{| None} syntax to indicate that a value may be temporarily unavailable, such as a robot position before its marker is detected. Install the dependencies with:
\begin{lstlisting}[language=bash]
python -m pip install numpy opencv-contrib-python roboid
\end{lstlisting}
The controller imports the robot interface with \code{from roboid import *}. RoboID is distributed through its maintainer's Python package page~\citep{roboidapi}. Matplotlib is used only for the optional error plot described in Section~\ref{sec:evaluation}; it is not part of the installation command above.

\vspace{-1pt}
\subsection{Controller settings}
\vspace{-1pt}
\label{sec:settings}
The example uses the following values in \code{settings.py}:
\begin{lstlisting}
target_ids = [0, 1, 2, 3]
ROBOTS_N = 4
SHAPE_R = 100
KP_DISTANCE = 0.7
KP_ANGLE = 0.5
DISTANCE_THRESHOLD = 10  # Control-vector magnitude, pixels
ANGLE_THRESHOLD = 30    # Degrees
OBS_THRESHOLD = 70
OBS_SPEED = 50
\end{lstlisting}
Here, \code{SHAPE_R} is the formation radius in \emph{image pixels}. The gains determine turning and forward commands, while the thresholds determine when those commands or the proximity response are selected. Their roles are explained in Section~\ref{sec:motion}.

The order of \code{target_ids} defines the mapping between marker IDs and robot connections. In this example, marker ID 0 corresponds to robot index 0, ID 1 to index 1, ID 2 to index 2, and ID 3 to index 3. Marker IDs do not need to match robot indices, as long as the mapping remains consistent. Connect the physical robots in the specified order, and ensure that \code{ROBOTS_N} equals the number of entries in \code{target_ids}. Note that marker IDs are distinct from RoboID connection IDs. During initialization, \code{slot=i} assigns each robot to the formation slot corresponding to its index.

\section{Robot Pose Estimation}
\label{sec:pose}
\subsection{ArUco detection and corner ordering}
Create a detector using the dictionary selected when printing the markers:
\begin{lstlisting}
aruco_dict = cv2.aruco.getPredefinedDictionary(
    cv2.aruco.DICT_7X7_100
)
detector = cv2.aruco.ArucoDetector(aruco_dict)
corners, ids, rejected = detector.detectMarkers(frame)
\end{lstlisting}
The result \code{corners} contains four image coordinates for each detected marker, and \code{ids} contains the corresponding marker IDs. If no markers are identified, \code{ids} is \code{None}. The third result contains rejected marker candidates; it is not a Boolean success flag.

\begin{figure}[htbp]
\centering
\includegraphics[width=.9\textwidth]{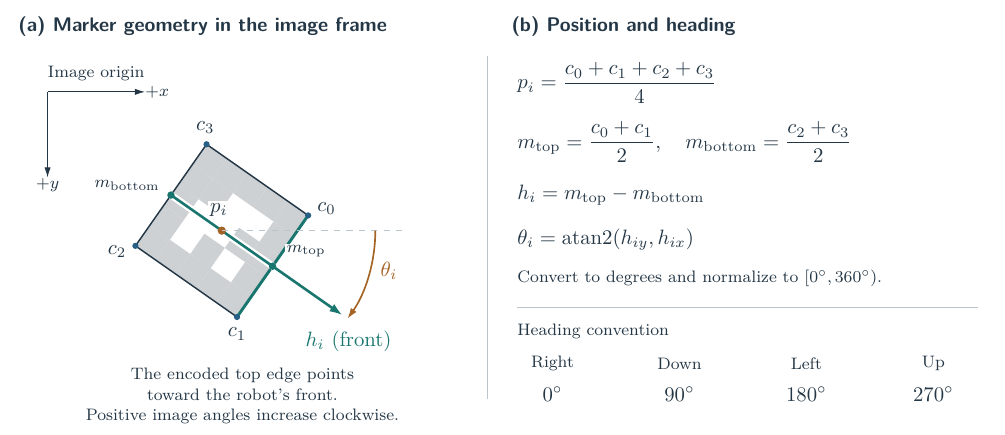}
\caption{Position and heading from the ordered marker corners. The image $x$-axis points right and the $y$-axis points down. The direction from the bottom-edge midpoint to the top-edge midpoint is the robot's forward direction when the marker is mounted as specified. The marker pattern is schematic, not a printable ArUco marker.}
\label{fig:pose}
\end{figure}

OpenCV orders the corners as top-left, top-right, bottom-right, and bottom-left \emph{relative to the marker's encoded orientation}~\citep{opencv2026aruco}. Denote them by $c_0,c_1,c_2,c_3$, with each $c_k$ a two-dimensional image coordinate. Figure~\ref{fig:pose} shows the geometry used below.

\subsection{Finding robot position}
The robot position is estimated by averaging the four marker corners:
\begin{equation}
    p_i=\begin{bmatrix}x_i\\y_i\end{bmatrix}
       =\frac{c_0+c_1+c_2+c_3}{4}.
    \label{eq:position}
\end{equation}
The averaging is applied separately to the horizontal and vertical coordinates. In Python:
\begin{lstlisting}
marker_corners = corners[index][0]
center = marker_corners.mean(axis=0)
x = center[0]
y = center[1]
\end{lstlisting}
Here, \code{index} selects one detected marker from the \code{corners} output. It is the marker's position in the returned list, not its encoded ID. Positions remain in camera pixel coordinates: the origin is at the top-left of the image, $x$ increases to the right, and $y$ increases downward. The same convention must be used in all subsequent calculations.

\subsection{Finding robot orientation}
Compute the midpoints of the top and bottom marker edges:
\begin{equation}
    m_{\mathrm{top}}=\frac{c_0+c_1}{2},\qquad
    m_{\mathrm{bottom}}=\frac{c_2+c_3}{2}.
\end{equation}
The forward vector and heading are
\begin{equation}
    h_i=m_{\mathrm{top}}-m_{\mathrm{bottom}},\qquad
    \theta_i=\operatorname{atan2}(h_{iy},h_{ix}).
    \label{eq:heading}
\end{equation}
The implementation converts the angle to degrees and normalizes it to $[0^\circ,360^\circ)$:
\begin{lstlisting}
topLeft, topRight, bottomRight, bottomLeft = marker_corners
top_center = (topLeft + topRight) / 2
bottom_center = (bottomLeft + bottomRight) / 2
heading_vector = top_center - bottom_center
theta = math.degrees(
    math.atan2(heading_vector[1], heading_vector[0])
)
theta %= 360
\end{lstlisting}
Because image $y$ increases downward, positive angles increase clockwise. The pose can be described as $[x_i,y_i,\theta_i]$, although the \code{RobotAgent} object stores position and orientation in separate fields.

\subsection{Updating poses safely}
The \code{update_poses()} function resets every robot's position and orientation before processing a new frame. A robot receives a valid pose only if its marker is detected in that frame.

The function belongs in \code{utils/camera_supervisor.py} and uses the detector initialized above:
\begin{lstlisting}
def update_poses(ROBOTS, frame):
    for robot in ROBOTS.values():
        robot.position = None
        robot.orientation = None
    corners, ids, _ = detector.detectMarkers(frame)
    if ids is None:
        return
    for marker_corner, marker_id in zip(corners, ids.flatten()):
        marker_id = int(marker_id)
        if marker_id not in ROBOTS:
            continue
        c = marker_corner.reshape((4, 2))
        topLeft, topRight, bottomRight, bottomLeft = c
        heading = ((topLeft + topRight)
                   - (bottomLeft + bottomRight)) / 2
        ROBOTS[marker_id].position = c.mean(axis=0)
        ROBOTS[marker_id].orientation = (
            math.degrees(math.atan2(heading[1], heading[0])) % 360
        )
\end{lstlisting}
The main loop stops the team when a required pose is missing. Without this reset, an early \code{return} when no markers are found would leave the previous frame's poses in place.

\textbf{Optional visual check.} To draw the detected marker outlines and IDs, add:
\begin{lstlisting}
if ids is not None:
    cv2.aruco.drawDetectedMarkers(frame, corners, ids)
\end{lstlisting}

\begin{figure}[htbp]

\centering
\includegraphics[width=.98\textwidth]{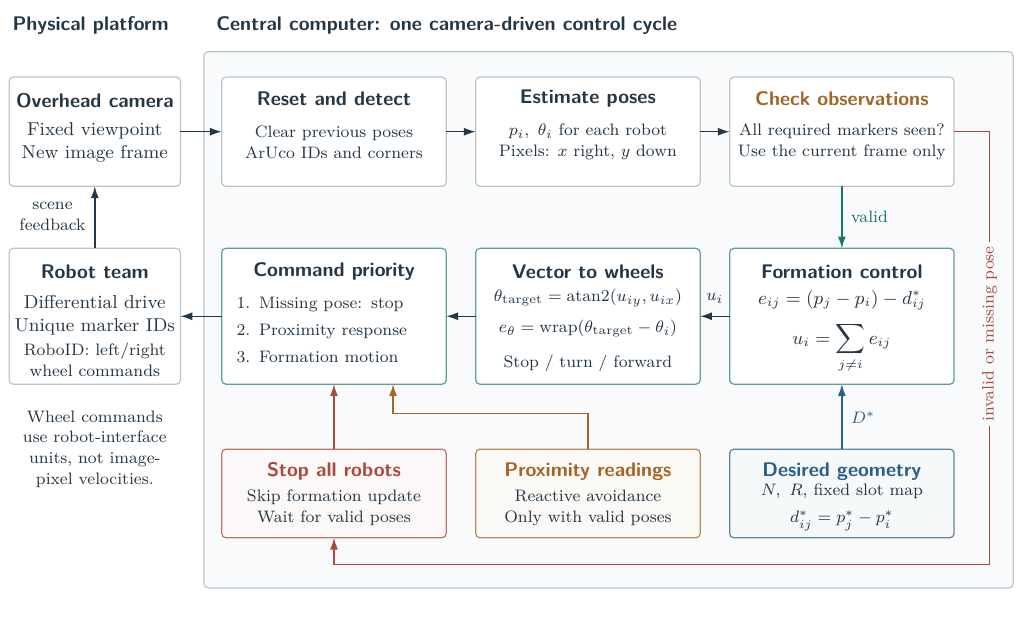}
\vspace{-5pt}
\caption{Frame-by-frame control flow. The host computer updates marker-based poses and stops all robots if a required pose is missing. Otherwise, it computes formation-control vectors and selects wheel commands, with the proximity response taking priority over formation motion. The pairwise error is current displacement minus desired displacement, $e_{ij}=(p_j-p_i)-d_{ij}^*$.}
\label{fig:architecture}
\end{figure}
\FloatBarrier

\vspace{-1pt}
\section{Formation Geometry and Control Vectors}
\vspace{-1pt}
\label{sec:formation}
The formation is defined in two steps. First, place fixed slots on a circle. Second, calculate the displacement from every slot to every other slot. The controller then compares these desired displacements with the measured robot-to-robot displacements.

\vspace{-1pt}
\subsection{Circular formation}
\begin{figure}[bp]
\centering
\includegraphics[width=0.95\textwidth]{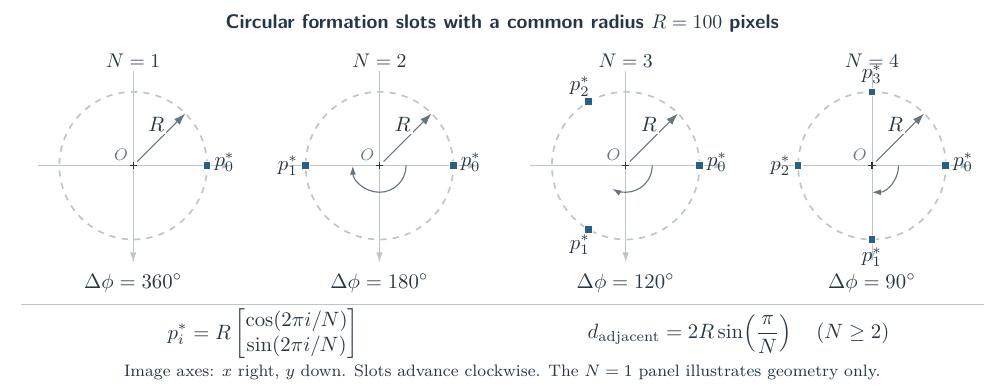}
\caption{Fixed slots on circles of equal radius, $R=100$ pixels. Identical axis bounds and aspect ratios give all four circles the same displayed diameter. Slot coordinates follow Eq.~\eqref{eq:slots}, using image axes with $y$ increasing downward. For four robots, the slots form a square.}
\vspace{-5pt}
\label{fig:slots}
\end{figure}

\vspace{-1pt}
For $N$ robots and radius $R$, consecutive slots are separated by $2\pi/N$ radians. Slot $i$ has desired coordinates
\begin{equation}
    p_i^*=\begin{bmatrix}
       R\cos(2\pi i/N)\\
       R\sin(2\pi i/N)
    \end{bmatrix},\qquad i=0,\ldots,N-1.
    \label{eq:slots}
\end{equation}
The collection of these positions is $P^*$ (``P star''). The configuration uses \code{ROBOTS_N} for $N$ and \code{SHAPE_R} for $R$. The corresponding function is
\begin{lstlisting}
def p_star():
    desired_positions = np.zeros((settings.ROBOTS_N, 2))
    for k in range(settings.ROBOTS_N):
        th = 2 * np.pi * k / settings.ROBOTS_N
        desired_positions[k] = [
            settings.SHAPE_R * np.cos(th),
            settings.SHAPE_R * np.sin(th),
        ]
    return desired_positions
\end{lstlisting}
Figure~\ref{fig:slots} shows the same construction for different robot counts. All four circles use the same radius and display scale. The $N=1$ panel illustrates slot geometry; the multi-robot tutorial uses two or more robots.

The slot coordinates are defined relative to $(0,0)$; they describe the formation shape, not a fixed destination in the camera image. The distance between adjacent slots is
\begin{equation}
    d_{\mathrm{adjacent}}=2R\sin\left(\frac{\pi}{N}\right).
    \label{eq:spacing}
\end{equation}
For \code{ROBOTS_N = 4} and \code{SHAPE_R = 100}, the adjacent center-to-center spacing is approximately 141.4 pixels. At a fixed radius, increasing the number of robots reduces the adjacent spacing. Select a radius that leaves room for the robots' physical size and movement.

\subsection{Desired pairwise displacements}
Rather than commanding robot $i$ to move to the image coordinate $p_i^*$, define the desired vector from its slot to slot $j$:
\begin{equation}
    d_{ij}^*=p_j^*-p_i^*.
    \label{eq:desired}
\end{equation}
The collection of these vectors is $D^*$ (``D star''). Each vector specifies both direction and distance. For example, $d_{ij}^*=(100,0)$ means that robot $j$ should be 100 pixels to the right of robot $i$. That relationship does not change when the whole formation is moved to another part of the camera image.

\begin{figure}[htbp]
\centering
\includegraphics[width=\textwidth]{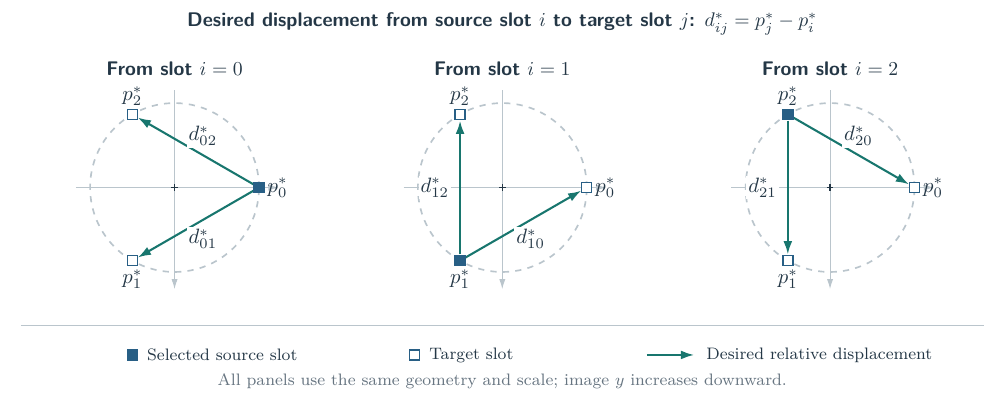}
\caption{Desired pairwise displacement vectors in a three-robot circular formation. Each arrow is calculated as target-slot position minus source-slot position, $d_{ij}^*=p_j^*-p_i^*$. The three panels use the same geometry and scale.}
\label{fig:displacements}
\end{figure}

Figure~\ref{fig:displacements} illustrates the desired vectors for three robots. In each panel, the filled square is the source slot, and the arrows point toward the other slots.

The function \code{d_star()} builds an array with shape $(N,N,2)$, using robot indices for array entries and slots for desired positions:
\begin{lstlisting}
def d_star(ROBOTS, desired_positions):
    desired_offset = np.zeros(
        (settings.ROBOTS_N, settings.ROBOTS_N, 2)
    )
    for robot_a in ROBOTS.values():
        for robot_b in ROBOTS.values():
            desired_offset[robot_a.robot_i, robot_b.robot_i] = (
                desired_positions[robot_b.slot]
                - desired_positions[robot_a.slot]
            )
    return desired_offset
\end{lstlisting}

\subsection{Formation error and control vector}
The measured displacement from robot $i$ to robot $j$ is
\begin{equation}
    d_{ij}=p_j-p_i.
\end{equation}
Subtract the desired displacement to obtain the pairwise error:
\begin{equation}
    e_{ij}=d_{ij}-d_{ij}^*=(p_j-p_i)-d_{ij}^*.
    \label{eq:error}
\end{equation}
If the current displacement equals the desired displacement, the error is zero. The subtraction order is important: the current displacement comes first.

For example, let $p_i=(100,100)$, $p_j=(70,160)$, and $d_{ij}^*=(-100,100)$. Then
\begin{align}
    d_{ij}&=(-30,60),\\
    e_{ij}&=(-30,60)-(-100,100)=(70,-40).
\end{align}
The horizontal and vertical components are calculated in the same image frame used for pose estimation. The positive horizontal and negative vertical components contribute a rightward and upward direction, respectively, to robot $i$'s control vector. Its final movement direction also depends on the errors relative to the other robots.

Each robot's control vector is the sum of its pairwise displacement errors:
\begin{equation}
    u_i=\sum_{\substack{j=0\\j\ne i}}^{N-1}e_{ij}
       =\sum_{\substack{j=0\\j\ne i}}^{N-1}\big[(p_j-p_i)-d_{ij}^*\big].
    \label{eq:control}
\end{equation}
The result $u_i=[u_{ix},u_{iy}]$ gives a desired movement direction and a magnitude used by the wheel-command routine. The sum excludes $j=i$ because a robot has zero displacement relative to itself. It is a sum of \emph{displacement errors}, not a sum of uncorrected robot-to-robot distances.

NumPy broadcasting calculates pairwise displacements for all robots without explicitly looping over each pair. The \code{None} entries introduce new array dimensions, allowing NumPy to subtract each robot's position from every other robot's position:
\begin{lstlisting}
actual_offset = positions[None, :, :] - positions[:, None, :]
displacement_error = actual_offset - desired_offset
u = np.sum(displacement_error, axis=1)
\end{lstlisting}
Here, \code{actual_offset[i, j]} equals $p_j-p_i$. The displacement arrays have shape $(N,N,2)$, and \code{u} has shape $(N,2)$, one movement vector per robot.

The use of pairwise differences connects this calculation conceptually to consensus-based coordination; Olfati-Saber et al.~\citep{olfatisaber2007consensus} provide an overview. Analyses of switching topologies and communication delays~\citep{olfatisaber2004consensus}, and of dynamically changing interaction topologies~\citep{ren2005consensus}, provide further theoretical background. Those results concern their respective models and assumptions; they do not establish convergence or delay tolerance for this camera-based turn-then-drive implementation.

\section{Differential-Drive Motion Control}
\label{sec:motion}
A differential-drive robot cannot directly execute an arbitrary planar displacement. It changes heading and moves forward by controlling its two wheels. This motion constraint is standard in mobile robotics~\citep{siegwart2011mobile}; we use a turn-then-drive rule.

\subsection{Target heading and angle error}
The target direction of $u_i$ is
\begin{equation}
    \theta_{\mathrm{target}}=\operatorname{atan2}(u_{iy},u_{ix}).
\end{equation}
Convert this angle to degrees, subtract the measured heading, and wrap the result to $[-180^\circ,180^\circ)$:
\begin{equation}
    e_\theta=((\theta_{\mathrm{target}}-\theta_i+180)\bmod360)-180.
    \label{eq:angleerror}
\end{equation}
In Python:
\begin{lstlisting}
target_angle = math.degrees(math.atan2(uy, ux))
angle_error = target_angle - theta
angle_error = (angle_error + 180) % 360 - 180
\end{lstlisting}
Both headings use the downward-$y$ image convention and degrees. Wrapping selects the smaller angular difference instead of requesting an unnecessary long turn.

\subsection{Turning}
When $|e_\theta|$ exceeds \code{ANGLE_THRESHOLD}, the robot turns in place by driving its wheels in opposite directions. We cap the turning command at 40:
\begin{equation}
    s_{\mathrm{turn}}=\min(K_{p,\theta}|e_\theta|,40),
\end{equation}
where $K_{p,\theta}$ corresponds to \code{KP_ANGLE}. The wheel commands are
\begin{lstlisting}
turn_speed = min(settings.KP_ANGLE * abs(angle_error), 40)
if angle_error > 0:
    robot.hamster.wheels(turn_speed, -turn_speed)
else:
    robot.hamster.wheels(-turn_speed, turn_speed)
\end{lstlisting}
For another robot platform, check its wheel-command signs rather than changing the formation-error equations.

\subsection{Forward motion and stopping}
The movement-vector magnitude is
\begin{equation}
    \norm{u_i}=\sqrt{u_{ix}^2+u_{iy}^2}.
\end{equation}
When this magnitude is at or below \code{DISTANCE_THRESHOLD}, the formation-control branch stops the robot. Otherwise, once the heading error is within \code{ANGLE_THRESHOLD}, both wheels receive the same forward command:
\begin{equation}
    s_{\mathrm{forward}}=\min(K_{p,d}\norm{u_i},40),
\end{equation}
where $K_{p,d}$ corresponds to \code{KP_DISTANCE}.
\begin{lstlisting}
distance = np.linalg.norm(u)
if distance <= settings.DISTANCE_THRESHOLD:
    robot.hamster.wheels(0, 0)
elif abs(angle_error) <= settings.ANGLE_THRESHOLD:
    forward_speed = min(settings.KP_DISTANCE * distance, 40)
    robot.hamster.wheels(forward_speed, forward_speed)
\end{lstlisting}
\textbf{Meaning of the stopping threshold.} Despite its name, \code{DISTANCE_THRESHOLD} is applied to the control-vector magnitude. It is not a direct measurement of inter-robot clearance, and it does not state that every pairwise error is below the same value. The normal turning and forward command caps are both 40; these are separate from \code{OBS_SPEED}.

\subsection{Reactive obstacle avoidance}
\label{sec:avoidance}
Before the formation-motion decisions above, \code{apply_control()} checks the left and right proximity sensors. Readings above \code{OBS_THRESHOLD} trigger the following reactive response:
\begin{lstlisting}
left = robot.hamster.left_proximity()
right = robot.hamster.right_proximity()
if left > settings.OBS_THRESHOLD and right > settings.OBS_THRESHOLD:
    robot.hamster.wheels(-settings.OBS_SPEED, -settings.OBS_SPEED // 2)
    return
if left > settings.OBS_THRESHOLD:
    robot.hamster.wheels(settings.OBS_SPEED, -settings.OBS_SPEED)
    return
if right > settings.OBS_THRESHOLD:
    robot.hamster.wheels(-settings.OBS_SPEED, settings.OBS_SPEED)
    return
\end{lstlisting}
The response temporarily overrides formation motion. It is basic reactive obstacle avoidance, not collision-free path planning. Once the sensor readings no longer trigger the response, the robot resumes the formation-motion decisions. The missing-marker stop remains outside this routine in the main loop.

\subsection{Tuning the controller}
\code{SHAPE_R} controls target spacing, \code{DISTANCE_THRESHOLD} controls the stopping condition for $\norm{u_i}$, and \code{ANGLE_THRESHOLD} controls how closely a robot must face its movement direction before moving forward. A smaller angle threshold requires closer alignment but can cause repeated turns. Begin with low wheel speeds; higher speeds can produce overshooting, collisions, or wheel slippage. Near the formation, reducing speed or increasing the stopping threshold can reduce oscillations.

\section{Main Control Loop}
\label{sec:loop}
The complete system repeatedly updates robot poses, calculates formation-control vectors, and sends wheel commands. The following excerpts show the main components of \code{main.py}.

\subsection{Robot initialization}
The \code{RobotAgent} class stores a marker ID, array index, robot connection, fixed formation slot, and the current position and orientation:
\begin{lstlisting}
from dataclasses import dataclass
import numpy as np
from roboid import *

@dataclass
class RobotAgent:
    marker_id: int
    robot_i: int
    hamster: HamsterS
    slot: int
    position: np.ndarray | None = None
    orientation: float | None = None

ROBOTS = {
    marker_id: RobotAgent(
        marker_id=marker_id,
        robot_i=i,
        hamster=HamsterS(i),
        slot=i,
    )
    for i, marker_id in enumerate(settings.target_ids)
}
\end{lstlisting}
The dictionary is keyed by marker ID. \code{robot_i} indexes the position and control-vector arrays, while \code{slot} selects a desired formation position. Position and orientation can be \code{None} before the first successful observation.

The main program also imports \code{cv2}, \code{settings}, \code{update_poses}, \code{p_star}, \code{d_star}, and \code{apply_control} from their respective modules in Table~\ref{tab:files}.

\subsection{Image capture and the missing-marker branch}
The program opens camera index 0 and requests 30 frames per second:
\begin{lstlisting}
cap = cv2.VideoCapture(0)
cap.set(cv2.CAP_PROP_FPS, 30)
\end{lstlisting}
The actual frame rate depends on the camera and computer. Each iteration calls \code{ret, frame = cap.read()}. Here, \code{ret} indicates whether the frame was read successfully. A failed read exits the loop. After a successful read, \code{update_poses(ROBOTS, frame)} updates the observations.

If any required position is missing, the loop stops every robot, displays the camera frame, checks whether the user pressed \code{q}, and otherwise skips the formation update:
\begin{lstlisting}
if any(robot.position is None for robot in ROBOTS.values()):
    for robot in ROBOTS.values():
        robot.hamster.wheels(0, 0)
    cv2.imshow("Camera", frame)
    if cv2.waitKey(1) & 0xFF == ord("q"):
        break
    continue
\end{lstlisting}
This behavior depends on clearing poses before processing each frame, as explained in Section~\ref{sec:pose}.

\subsection{Formation calculation and actuation}
For a frame with all required poses, populate the position array, construct the desired displacements, calculate the errors, and send each robot its own vector:
\begin{lstlisting}
positions = np.zeros((settings.ROBOTS_N, 2))
for robot in ROBOTS.values():
    positions[robot.robot_i] = robot.position

desired_positions = p_star()
desired_offset = d_star(ROBOTS, desired_positions)
actual_offset = positions[None, :, :] - positions[:, None, :]
displacement_error = actual_offset - desired_offset
u = np.sum(displacement_error, axis=1)

for robot in ROBOTS.values():
    apply_control(robot, u[robot.robot_i], robot.orientation)
\end{lstlisting}
These calculations implement Eqs.~\eqref{eq:slots}--\eqref{eq:control}. The wheel-command routine implements the turning, forward-motion, stopping, and proximity decisions in Section~\ref{sec:motion}.

\subsection{Display and cleanup}
The main loop displays the camera frame in the OpenCV window and uses \code{cv2.waitKey(1)} to detect the quit key. The camera image is not automatically annotated unless marker drawing or the optional overlay is enabled.

Place the loop inside a \code{try} block and use a \code{finally} block to stop the wheels and release the camera and display resources:
\begin{lstlisting}
finally:
    for robot in ROBOTS.values():
        robot.hamster.wheels(0, 0)
    cap.release()
    cv2.destroyAllWindows()
\end{lstlisting}

\section{Running the System}
\label{sec:run}
\paragraph{Step 1: Prepare the robots.}
Attach one unique ArUco marker to each robot. Turn on and connect the HamsterS robots individually in the order corresponding to \code{target_ids}, then place them inside the camera view.

\paragraph{Step 2: Position and check the camera.}
Keep the camera stationary above the operating area. Confirm that all markers are visible, face the camera, and appear in focus under adequate lighting. Before starting formation control, verify that each detected marker ID matches its robot, the reported heading points toward the robot's front, and a short turn command rotates in the intended direction.

\needspace{7\baselineskip}
\paragraph{Step 3: Start the program.}
Run:
\begin{lstlisting}[language=bash]
python main.py
\end{lstlisting}
The camera window should display the live image. Marker outlines, IDs, and vector overlays appear only when the relevant optional drawing calls are enabled. The robots should turn toward their control vectors and move to reduce their formation errors.

\paragraph{Step 4: Stop the system.}
Press \code{q} in the OpenCV window. The cleanup block stops every robot before closing the camera and display.

\section{Expected Behavior and Error Measurement}
\label{sec:evaluation}
The robots can begin at different positions and orientations within the operating area. As the loop runs, the camera estimates their poses, the controller computes displacement errors, and the robots turn and drive to reduce those errors. The formation-motion branch stops a robot when its control-vector magnitude is within the threshold, unless the proximity response takes precedence.

With four robots, the expected arrangement is a square around a common formation center, not a single shared pixel. Robot size, camera noise, wheel-speed differences, and the stopping threshold can cause the physical formation to differ from the ideal geometry. Figure~\ref{fig:sequence} shows one experimental run as a qualitative illustration.

\subsection{Measuring formation error}
We measure the mean pairwise formation error as
\begin{equation}
    E(t)=\frac{1}{N(N-1)}
    \sum_{\substack{i,j=0 \\ i\ne j}}^{N-1}
    \norm{e_{ij}(t)}.
    \label{eq:metric}
\end{equation}
There are $N(N-1)$ ordered pairs of distinct robots. The metric averages the magnitude of the displacement mismatch over those pairs; a smaller value means a closer match to the desired relative arrangement.

The function \code{global_error()} in \code{utils/img_overlay.py} calculates this metric, while \code{draw_overlay()} provides optional visualization. To record the error, initialize an elapsed-time clock and two lists before the main loop:
\begin{lstlisting}
import time
from utils.img_overlay import global_error, draw_overlay

error_times, errors = [], []
start_time = time.perf_counter()
\end{lstlisting}
After calculating the displacement errors and control vectors in a valid frame, add:
\begin{lstlisting}
error = global_error(positions, desired_offset)
error_times.append(time.perf_counter() - start_time)
errors.append(error)
draw_overlay(frame, ROBOTS, u, error)  # Before cv2.imshow(...)
\end{lstlisting}
Only frames in which all required robots are detected contribute samples. With Matplotlib installed, plot the recorded values after the run:
\begin{lstlisting}
import matplotlib.pyplot as plt

plt.plot(error_times, errors)
plt.xlabel("Time (s)")
plt.ylabel("Mean formation error (px)")
plt.savefig("formation_error_over_time.png", dpi=200)
plt.close()
\end{lstlisting}
These lines are optional and are not part of the main loop by default. The measured trace can fluctuate because of motion and sensing noise.

\section{Troubleshooting}
\label{sec:troubleshooting}
Table~\ref{tab:troubleshooting} summarizes common issues and the corresponding checks. Use the same coordinate convention and angular units throughout; changing one sign in isolation can introduce a new inconsistency.

\begin{table}[htbp]
\centering\small
\caption{Common implementation issues and troubleshooting checks.}
\label{tab:troubleshooting}
\begin{tabularx}{\textwidth}{@{}p{.25\textwidth}Y@{}}
\toprule
Symptom & Checks and adjustments\\
\midrule
Camera does not open & Try another camera index, such as \code{cv2.VideoCapture(1)}, and check that another program is not using the camera.\\
No markers detected & Check the dictionary, marker IDs, lighting, focus, marker size, and visibility of the complete marker border.\\
Robot moves incorrectly & Check image coordinates, marker mounting direction, wheel signs, and the angular-error calculation. Both measured and target headings must use degrees and the same image frame.\\
Robot turns the wrong way & Check the platform's left- and right-wheel sign conventions. Adapt the wheel mapping in \code{apply_control()} rather than changing the formation-error equations.\\
Robots oscillate near the formation & Reduce wheel speed or increase the stopping threshold. Very tight heading tolerance can also produce repeated turning.\\
\bottomrule
\end{tabularx}
\end{table}
\FloatBarrier

\section{Limitations}
\label{sec:limitations}

Although this tutorial provides a practical introduction to multi-robot formation control, the implementation has several limitations. First, robot positions and formation dimensions are expressed in image pixel coordinates rather than physical units. Without camera calibration and image-to-world coordinate conversion, pixel distances do not necessarily correspond to uniform physical distances across the workspace. Pose estimation also depends on reliable ArUco marker detection and proper marker placement.

Second, the system uses a centralized architecture with a fixed overhead camera. All robot poses are estimated and control commands are computed on a single computer. Consequently, the system requires all robot markers to remain visible; if any required marker is lost, the robots stop until valid observations are restored. Furthermore, the formation geometry and robot-to-slot assignments are predefined, without dynamic reassignment or adaptation of the desired formation.

Finally, the controller uses a simple turn-then-drive strategy with fixed gains and thresholds. Its stopping condition is based on the magnitude of the combined formation-control vector, rather than directly enforcing individual pairwise error bounds. Although proximity sensors provide reactive obstacle avoidance, the implementation does not include collision-free path planning or explicit inter-robot separation constraints. Therefore, reducing formation error does not guarantee collision-free motion or convergence under all operating conditions.

\section{Conclusion}

This tutorial presented a step-by-step implementation of vision-based multi-robot formation control using differential-drive robots, an overhead camera, and Python. We introduced the key components of a multi-robot control system, from ArUco-based robot identification and pose estimation to desired formation geometry, relative displacement errors, and differential-drive motion control. By integrating these components into a continuous perception-and-control loop, we demonstrated how mathematical formation-control principles can be translated into coordinated robot behavior.

A primary motivation for developing this tutorial is to help bridge the gap between theoretical concepts and practical implementation in robotics education. While mathematical models and control algorithms provide the foundation for understanding multi-robot coordination, implementing them on physical robots introduces additional challenges involving sensing, coordinate systems, actuation, and system integration. Through an accessible, hands-on example, we aim to help students not only understand how formation control works, but also gain the confidence and practical skills to implement, experiment with, and further explore multi-robot systems.

We hope this tutorial serves as a useful educational resource for students, educators, and researchers who wish to gain hands-on experience in robotics, particularly in multi-robot systems. One of our primary goals is to make practical multi-robot experimentation accessible without requiring expensive robotic platforms or specialized laboratory facilities. The HamsterS robots used in our implementation are relatively affordable, costing approximately US\$100 per robot at the time of writing. Moreover, the underlying formation-control principles can be adapted to other differential-drive robots by modifying the platform-specific sensing, communication, and actuation interfaces. We hope this accessibility encourages readers to build their own multi-robot systems, experiment with real robots, and deepen their understanding of robotics through practical implementation.

For readers who do not have access to physical robots, the Robotarium at the Georgia Institute of Technology provides an alternative platform for remotely conducting experiments with real multi-robot systems~\citep{pickem2017robotarium,wilson2020robotarium}. Although adapting the implementation to the Robotarium requires using its supported interfaces, it offers an opportunity to explore multi-robot coordination without owning robotic hardware. The accompanying GitHub repository provides the Python implementation, setup instructions, and demonstrations to support independent learning and experimentation. Ultimately, we hope this tutorial helps bridge the gap between theory and practice, inspires readers to explore more advanced problems in multi-robot systems, and encourages them to develop their own ideas and research in robotics.

\section*{Acknowledgments}
The authors would like to thank the students enrolled in Multi-Robot Systems (CSCI-B 490/590 and ENGR-E 399/599) at Indiana University Bloomington in Spring 2026 for their insightful discussions and valuable feedback. We also thank SMART Lab alumni Dr. Vishnunandan Venkatesh (Astemo Americas, Inc.) and Prof. Wonse Jo (Incheon National University) for their contributions to developing the Hamster robot control system and SMART Table used in SMART Lab technical seminars. 

Generative AI tools were used to assist with language editing and figure preparation. All AI-assisted content was carefully reviewed and verified by the authors.

\begingroup
\renewcommand{\bibfont}{\small}
\setlength{\bibsep}{2pt}
\interlinepenalty=10000
\bibliographystyle{IEEEtran}
\bibliography{references}
\endgroup
\end{document}